\documentclass[10pt,twocolumn,letterpaper]{article}

\usepackage[pagenumbers]{cvpr} 

\usepackage{microtype}

\renewcommand{\paragraph}[1]{\vspace{.3em}\noindent\textbf{#1}}

\definecolor{cvprblue}{rgb}{0.21,0.49,0.74}
\usepackage[pagebackref,breaklinks,colorlinks,allcolors=cvprblue]{hyperref}

\def\paperID{9} 
\def\confName{CVPR}
\def\confYear{2026}

\title{Optimizing Latent Representations for Robust Building Damage Assessment Onboard Earth Observation Satellites}

\author{Thomas Goudemant, Benjamin Francesconi\\
Institut de Recherche Technologique Saint Exupéry\\
{\tt\small\{thomas.goudemant, benjamin.francesconi\}@irt-saintexupery.com}
}
\begin{document}
\maketitle


\begin{abstract}
Rapid identification of damaged buildings after natural disasters or on war areas is crucial to support emergency response and prioritize interventions. Earth Observation constellations provide timely, large-scale coverage, but actionable information is often delayed by data downlink constraints, on-ground processing, and human interpretation. Reducing this latency is essential to improve decision-making responsiveness.
In this work, we propose an original AI-based system that enables object-level building damage assessment (localization and damage classification) directly on-board satellites from pre-disaster and post-disaster high-resolution optical imagery. Available pre-disaster images are encoded on ground into compact latent representations, transmitted to the satellite, and compared on-board with newly acquired post-event observations. Leveraging AI interpretation capabilities and increasing processing capabilities on-board satellites, the proposed design enables processing directly at the data source, reducing the amount of information to be downlinked while preserving task-relevant content and improving overall system responsivity.
We explore the design space through a systematic benchmark of on-board-compatible variants, analyzing the impact of siamese processing, cross-attention, latent-space compression, and robustness-oriented data augmentation. Experiments on xBD dataset demonstrate reliable and robust damage assessment under misalignment, with minimal performance degradation under strong compression.
\end{abstract}    
\section{Introduction}
\label{sec:intro}

Timely assessment of building damage after natural disasters or armed conflicts is critical to support emergency response, allocate limited resources, and prioritize field interventions. In the first hours following a catastrophic event, situational awareness over affected urban areas is essential, yet ground-based inspections are often slow, hazardous, and logistically constrained \cite{gupta2019xbddatasetassessingbuilding}. Rapidly localizing buildings and estimating their damage level therefore plays a central role in operational disaster management.

High-resolution (HR) Earth Observation (EO) satellites provide large-scale, repeatable coverage of impacted regions and have become a key source of post-disaster information. In parallel, advances in deep learning have significantly improved automated building damage assessment from satellite imagery \cite{10645210,gupta2019xbddatasetassessingbuilding}, enabling localization and classification of damaged structures over wide areas.

Most high-performing methods formulate building damage assessment as a change detection or bi-image problem, jointly exploiting pre- and post-disaster imagery to reduce ambiguities between newly destroyed and pre-existing structures \cite{rs12101670,ZHENG2021112636}. Although robust, such pipelines remain largely ground-centric, requiring full image downlink and offline processing before actionable products are generated \cite{rs15163963}, resulting in operational latency.

Recent advances in intelligent satellite systems and embedded AI accelerators enable moving part of the processing closer to the sensor \cite{9705087}. Onboard object detection has been demonstrated for several EO tasks \cite{rs15163963,phisat2}, confirming the feasibility of embedded deep learning in orbit. Extending this concept to bi-temporal building damage assessment, however, introduces additional challenges including limited uplink bandwidth for transmission of pre-disaster reference data, restricted onboard memory and compute resources, robustness to imperfect pre-/post-disaster image co-registration, and compatibility with raw or minimally processed data.

In this work, we propose a detection-based building damage assessment architecture from pre- and post-disaster HR optical EO imagery, specifically designed for embedded deployment. Our approach separates pre- and post-disaster processing while preserving the benefits of bi-temporal reasoning. Pre-disaster imagery is encoded on the ground into compact latent representations that are uplinked and stored on board. Newly acquired post-disaster images are processed directly in orbit and compared with the stored latent features to perform object-level damage assessment, substantially reducing communication requirements.

\paragraph{Contributions.} (i) We introduce an innovative ground/onboard architecture for bi-temporal building damage assessment that decouples pre- and post-event processing to overcome ground-to-space communication constraints.
(ii) We provide a systematic ablation study of design choices—including early fusion versus siamese processing, feature-level cross-attention, and latent-space compression.  
(iii) We analyze robustness and deployability under realistic operational conditions, including severe pre/post misregistration and bandwidth constraints in uplink and downlink volumes.

We next position our approach with respect to prior work in bi-temporal change detection, attention-based architectures, object-level damage assessment, and intelligent on-board EO systems.

\section{Related Work}
\label{sec:related}
To position our contribution, we review prior work along four complementary axes: 
(i) bi-temporal formulations for building damage assessment and siamese change detection paradigms, 
(ii) the evolution from dense pixel-level segmentation to object-level damage modeling, 
(iii) transformer-based strategies for multi-temporal feature interaction, 
and (iv) onboard and embedded change detection systems. 
This structured analysis clarifies how methodological advances in multi-temporal reasoning and object-level representation have progressed, while highlighting the remaining gap between high-performing ground-based approaches and operationally deployable embedded systems.

\paragraph{Bi-Temporal and Siamese Change Detection.} 
Building damage assessment from EO imagery is commonly formulated as a bi-temporal change detection problem \cite{ANDRESINI2023119123}. In disaster assessment scenarios, the use of paired pre- and post-event imagery is frequently adopted, as relying solely on post-disaster observations may cause confusion between damaged buildings and background structures \cite{rs12101670}. Existing methods are implemented either as cascade pipelines, where building localization and damage classification are performed in successive stages, or as unified end-to-end architectures. Cascade formulations often suffer from a knowledge gap between localization and classification stages, as the two components are optimized separately, limiting joint feature learning \cite{ZHENG2021112636}. To reduce this limitation, fully convolutional siamese networks, introduced for remote sensing change detection by Daudt et al. \cite{DaudtSiamese}, process pre- and post-event images through weight-sharing encoders, enabling temporally consistent feature extraction and direct comparison in a shared representation space.

\paragraph{From Dense Segmentation to Object-Level Damage Detection.}
While bi-temporal architectures improve feature alignment and change identification, most building damage assessment methods are formulated as dense pixel-level segmentation tasks, as exemplified by the xBD benchmark \cite{gupta2019xbddatasetassessingbuilding}. However, pixel-wise modeling may lead to semantic inconsistencies at the building scale, especially when damage affects only part of a structure, resulting in mixed predictions within a single instance \cite{ZHENG2021112636}. 
Object-based approaches instead consider buildings as the fundamental unit of analysis rather than individual pixels, enabling more coherent semantic predictions and reducing noise and spurious detections \cite{rs14071552}. In this formulation, damage assessment can be cast as an object-level detection problem, where buildings are localized and assigned a single damage state. Although originally developed for mono-temporal scenarios, standard object detection frameworks such as SSD \cite{DBLP:journals/corr/LiuAESR15}, Faster R-CNN \cite{DBLP:journals/corr/RenHG015}, or YOLO \cite{DBLP:journals/corr/RedmonDGF15} provide suitable architectural backbones for such modeling when extended to multi-temporal inputs.

\paragraph{Transformer-Based Multi-Temporal Modeling.}
Transformer architectures based on self-attention have demonstrated strong ability to model long-range dependencies \cite{Vaswani2017}. Their adaptation to visual tasks through Vision Transformers confirmed the effectiveness of global attention for capturing rich spatial context \cite{Dosovitskiy2020}. Recent works have incorporated transformer-based designs into bi-temporal remote sensing pipelines, including siamese transformer architectures that leverage attention mechanisms to enhance feature interaction between pre- and post-event imagery \cite{Mohammadian_2023}, as well as multi-scale attention frameworks specifically designed for building damage assessment \cite{math10111898}. These approaches highlight the importance of global contextual reasoning for multi-temporal modeling, particularly in complex urban scenes where long-range spatial dependencies can provide critical cues for damage interpretation.

\paragraph{Onboard Deployment and Operational Constraints.}
The integration of embedded AI within EO satellites enables onboard processing to reduce latency and communication load \cite{9705087}. Deep learning models have been deployed in orbit, including convolutional change detection systems \cite{10148624} and object detection frameworks adapted to spaceborne hardware \cite{rs15163963,goudemant:hal-03881738}, while system-level demonstrators have explored reactive ground–space architectures \cite{francesconievent} that shorten the decision–action loop. However, most existing approaches address either mono-temporal detection or pixel-level change detection, without jointly considering bi-temporal reasoning, object-level damage modeling, and the practical constraints of embedded deployment. In realistic scenarios, coping with limited onboard memory and compute resources, uplink/downlink bandwidth restrictions, residual pre/post co-registration errors, and the handling of raw sensor data remain critical \cite{Dorise}. The explicit integration of pre-event reference information in embedded multi-temporal pipelines is still largely underexplored.

\paragraph{Summary and positioning.} Enabling efficient multi-temporal building detection under strict bandwidth, compute, and memory limits, as well as robustness to onboard raw data inaccuracies (e.g., geolocation uncertainty), remains an open operational challenge. Addressing this gap requires the development of new AI-based architectures capable of localizing and classifying individual damaged buildings from two or more temporally distinct images, while explicitly accounting for ground–space communication constraints and embedded deployment limitations. Our work makes a step in this direction. 

\section{Method}
\label{sec:method}

To address this gap, we introduce a ground/on-board architecture specifically designed for object-level building damage assessment under realistic operational constraints.

\subsection{Ground/On-board System Concept}
\label{sec:system_concept}

We consider an operational Earth Observation scenario in which rapid assessment of building damage is required to support post-disaster response. In this setting, algorithmic performance and system-level constraints must be addressed jointly.

From an algorithmic standpoint, we adopt a bi-temporal architecture taking as input pre-/post-disaster images, consistently with prior work~\cite{rs12101670} showing its superior performance over post-event analysis alone. From an operational standpoint, improving responsiveness requires moving inference closer to the sensor~\cite{9705087}, thereby shortening the decision loop and reducing downlink data volume with respect to systematic full-resolution imagery. However, this system-level shift introduces strict constraints: uplink bandwidth and onboard memory prevent transmitting complete pre-disaster reference scenes, while limited onboard geolocation accuracy induces residual misregistration between pre- and post-event observations.

To reconcile these constraints, pre-disaster scenes are encoded on the ground into compact latent representations that are uploaded and stored onboard. These representations are designed to (i) significantly reduce uplink volume and (ii) improve robustness to onboard data imperfections, particularly residual misregistration due to limited geolocation accuracy. When post-disaster imagery is acquired, the onboard module compares it with the stored latent features to perform object-level building damage assessment.

\begin{figure}[ht]
\centering
\includegraphics[width=.99\columnwidth]{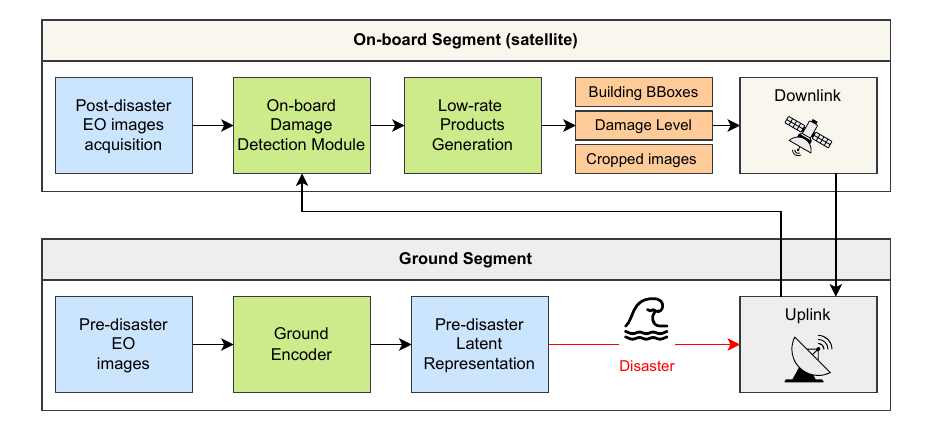}
\caption{Ground/on-board concept. Pre-disaster scenes are encoded on the ground into compact latents uplinked to the satellite. Post-disaster images are compared on-board to detect damage and generate low-rate products.}
\label{fig:pipeline}
\vspace{-6mm}
\end{figure}

The onboard module outputs concise products consisting of localized building instances (bounding boxes) and associated damage levels. Optionally, image crops over detected areas of interest can be generated. By transmitting only structured detection results and targeted patches rather than full scenes, the proposed concept substantially reduces downlink volume while preserving actionable information. The overall processing concept is illustrated in Fig.~\ref{fig:pipeline}.

\subsection{Base Detection Architecture}
\label{sec:base_arch}

\begin{figure*}[t]
  \centering
  \begin{subfigure}{0.48\linewidth}
    \centering
    \includegraphics[height=3.4cm]{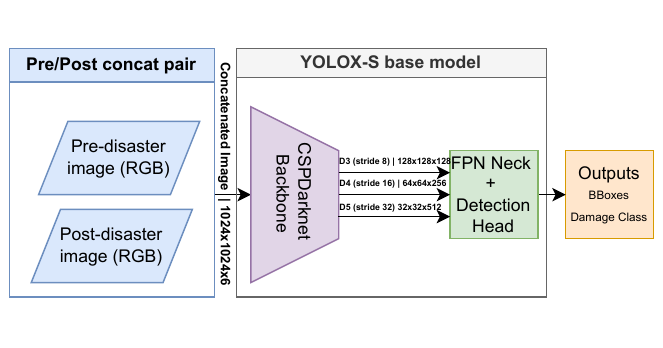}
    \caption{Baseline early-fusion architecture (6-channel input).}
    \label{fig:baseline}
  \end{subfigure}
  \hfill
  \begin{subfigure}{0.48\linewidth}
    \centering
    \includegraphics[height=3.4cm]{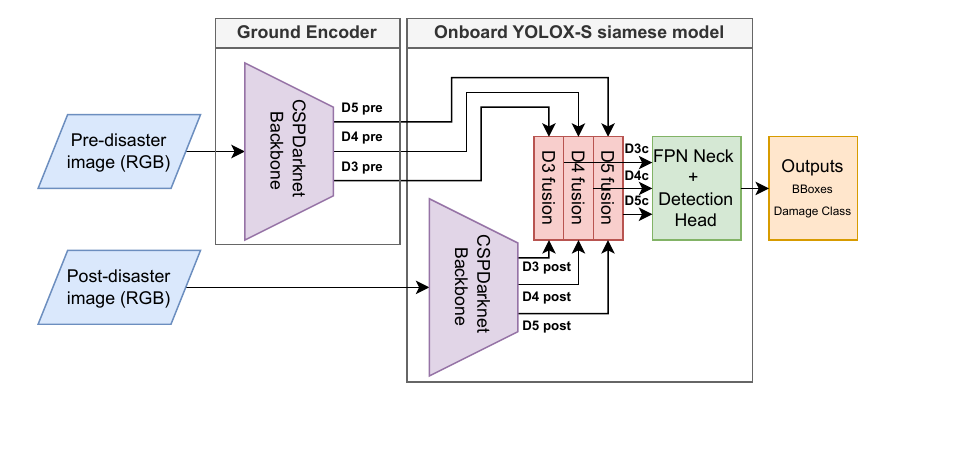}
    \caption{Siamese backbone with shared weights.}
    \label{fig:siamese}
  \end{subfigure}

  \begin{subfigure}{0.48\linewidth}
    \centering
    \includegraphics[height=3cm]{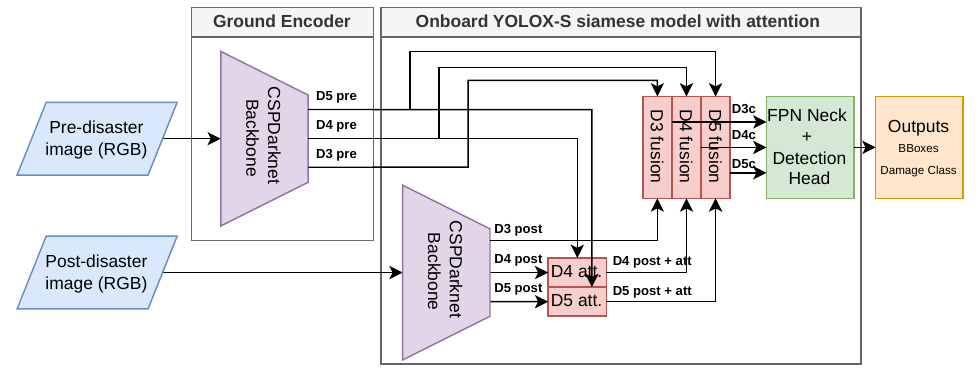}
    \caption{Cross-attention between pre- and post-disaster features.}
    \label{fig:attention}
  \end{subfigure}
  \hfill
  \begin{subfigure}{0.48\linewidth}
    \centering
    \includegraphics[height=3cm]{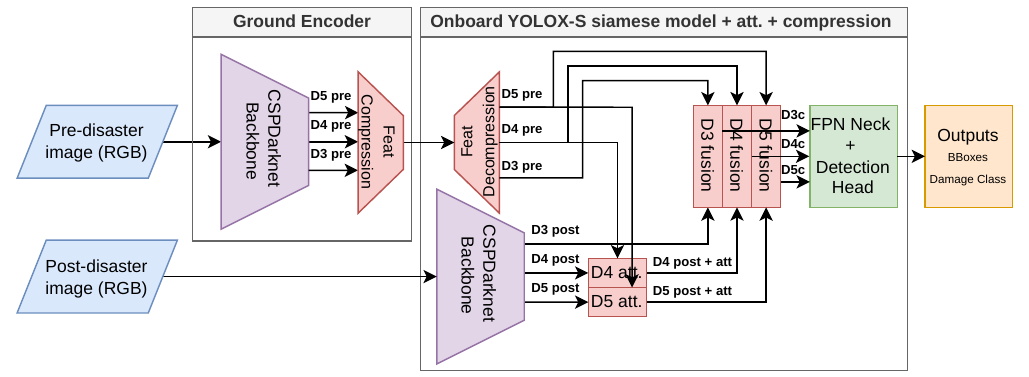}
    \caption{Compressed pre-disaster latent representations.}
    \label{fig:compression}
  \end{subfigure}
\caption{Overview of the baseline architecture and the proposed architectural variants. All variants share the same backbone, neck and head but differ in the processing of pre- and post-disaster latent features.}
\label{fig:architectures}
\vspace{-4mm}
\end{figure*}

Building damage assessment is formulated as an object detection task, where buildings are localized and assigned a discrete damage level. To satisfy latency, memory, and implementation constraints associated with onboard deployment, we adopt a single-stage detection architecture as the common backbone for all experiments.

Our base detector follows the standard design of modern single-stage detectors and consists of a convolutional backbone, a feature pyramid network (FPN), and a detection head. In practice, we instantiate this architecture using YOLOX-S~\cite{ge2021yolox}, selected for its favorable trade-off between detection accuracy and inference speed, as well as its compatibility with embedded processing targets~\cite{electronics11213473}. The detection framework itself is not a contribution of this work and serves as a controlled reference architecture.

The CSPDarknet backbone produces three intermediate feature maps (D3, D4, D5) at spatial strides of 8, 16, and 32 pixels. These multi-scale representations define the feature space in which pre- and post-disaster information is combined and compared. The FPN aggregates these features across scales, and a shared detection head predicts bounding boxes and damage classes at each level. The FPN and detection head remain architecturally unchanged across all variants, although all components are trained end-to-end.

As illustrated in Fig.~\ref{fig:baseline}, the baseline configuration performs early fusion by channel-wise concatenation of pre- and post-disaster images, resulting in a 6-channel input to the backbone. All architectural variants introduced in the next section—including siamese processing, cross-attention mechanisms, and latent-space compression—operate on the backbone feature maps (D3--D5) while preserving the same detection pipeline.

This early-fusion configuration assumes joint availability of pre- and post-disaster imagery at inference time, without accounting for high-bandwidth communication constraints. It therefore serves as a reference model for quantifying the impact of the proposed deployment-oriented modifications.

\subsection{Architectural Variants}
\label{sec:variants}
The following section introduces a set of architectural variants that progressively relax system constraints and enable controlled evaluation of on-board–compatible processing strategies.

\subsubsection{Siamese Processing and Feature Fusion}
\label{sec:siamese}
We extend the baseline detector with a siamese formulation inspired by bi-temporal change detection paradigms. Instead of fusing pre- and post-disaster imagery at the input level, the two images are processed independently by identical backbones with shared weights (Fig.~\ref{fig:siamese}). The shared-weight design ensures that features extracted on both images lie in a common representation space, enabling consistent comparison while keeping the number of parameters unchanged.

At each backbone stage (D3, D4, D5), feature maps from the pre- and post-disaster branches are fused prior to the FPN. For each stage $n \in \{3,4,5\}$, we concatenate the pre-disaster features, the post-disaster features, and their element-wise difference to form the fused representation $D_{nc}$. This formulation exposes both absolute contextual information and explicit change cues to the detection pipeline.

Beyond improved multi-temporal reasoning, this siamese structure naturally supports the ground/on-board separation introduced in Section~\ref{sec:system_concept}. Pre-disaster feature maps can be computed on the ground and reused onboard, enabling later extensions where only compact latent representations are uplinked.

\subsubsection{Cross-Attention between Pre and Post Features}
\label{sec:attention}
In operational onboard settings, residual spatial misregistration between pre- and post-disaster images is unavoidable due to acquisition geometry differences and limited onboard registration capabilities. Under such conditions, strictly local feature comparison may fail to retrieve relevant pre-disaster information at the correct spatial locations. To mitigate this issue, we introduce cross-attention mechanisms that enable spatially adaptive interaction between pre- and post-disaster feature maps.

\begin{figure}[ht]
\centering
\includegraphics[width=.8\columnwidth]{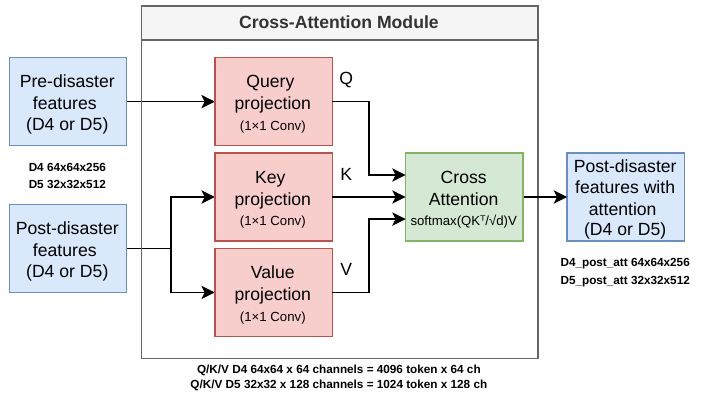}
\caption{Principle of cross-attention between pre- and post-disaster features. Pre-disaster features act as queries to retrieve relevant contextual information from post-disaster features (keys and values), enabling robust comparison under spatial misregistration before feature fusion.}
\label{fig:attention_principle}
\vspace{-2mm}
\end{figure}

We implement an asymmetric cross-attention module following the query–key–value formulation introduced by Vaswani et al.~\cite{Vaswani2017}. In our design, pre-disaster backbone features act as queries, while post-disaster features provide keys and values (Fig.~\ref{fig:attention_principle}). Query, key, and value embeddings are obtained through $1\times1$ convolutions applied to the backbone feature maps. The $1\times1$ projections also reduce the channel dimensionality by a factor of four, limiting the computational cost of the attention operation while preserving sufficient representational capacity. This mechanism allows each spatial location in the pre-disaster representation to attend over the entire post-disaster feature map and retrieve a weighted combination of features from potentially shifted locations, without requiring explicit geometric alignment.

The resulting attention output is used to refine the post-disaster representation prior to the feature fusion described in Sec.~\ref{sec:siamese}. The integration of the cross-attention module within the overall siamese ground/on-board architecture is illustrated in Fig.~\ref{fig:attention}. This refinement enables more robust comparison through subsequent concatenation and differencing.

To balance robustness and computational efficiency, cross-attention is applied only to higher-level feature maps (D4 and D5), where spatial resolution is lower and semantic abstraction is higher. The finest-scale features (D3) are excluded in the default configuration, as they encode low-level spatial details and would incur a significantly higher computational cost due to the quadratic scaling of attention with spatial resolution. The impact of applying attention at different feature levels is analyzed experimentally in Sec.~\ref{sec:experiments}.

\subsubsection{Compressed Pre-Disaster Latent Representations}
\label{sec:compression}

In the proposed ground/on-board system, pre-disaster information is represented by multi-scale backbone feature maps computed on the ground and uplinked for later comparison with post-disaster observations. Although this latent representation enables direct feature-level comparison onboard, its raw size can exceed that of the original input image. For a $1024 \times 1024$ input, the uncompressed feature maps (D3, D4, D5) amount to approximately 3.5\,MB, compared to 3.0\,MB for the RGB image, making uplink compression a critical system-level requirement.

To address this issue, we apply channel-wise compression prior to uplink. At each scale, the backbone features are projected into a lower-dimensional space through a lightweight $1\times1$ convolution that reduces the number of channels by a factor $r$, with $r \in \{8, 64\}$ in our experiments. This operation preserves the original spatial resolution of each feature map while significantly reducing memory footprint. The compressed feature maps are uplinked and, upon reception onboard, expanded through a symmetric lightweight projection trained jointly with the overall detection network to match the channel dimensionality expected by the fusion pipeline (Fig.~\ref{fig:compression}).

In addition to channel reduction, we explore configurations that remove the finest-resolution feature level (D3) from both pre- and post-disaster branches. Since D3 primarily captures fine-grained spatial details (e.g., edges and textures) and contributes significantly to memory and computation cost, its removal further reduces uplink volume and onboard processing requirements. The quantitative impact of this design choice is analyzed in Sec.~\ref{sec:quantitative}.

Tab.~\ref{tab:latent_compression} summarizes the resulting memory footprint of the pre-disaster latent representation under different compression configurations, reported as equivalent int8 sizes to reflect quantized onboard inference targets. The quantitative effects of these compression strategies on detection accuracy and robustness are evaluated in Sec.~\ref{sec:quantitative}.

\begin{table}[t]
\centering
\caption{Size of pre-disaster latent representations under different channel compression configurations for an input image of size $1024\times1024\times3$. Sizes are reported as equivalent int8 memory footprint.}
\vspace{-2mm}
\label{tab:latent_compression}
\footnotesize
\setlength{\tabcolsep}{4pt}
\begin{tabular}{lcccc}
\toprule
\textbf{Configuration} &
\textbf{D3} &
\textbf{D4} &
\textbf{D5} &
\textbf{Size (MB)} \\
\midrule
Input image ($1024{\times}1024{\times}3$) & -- & -- & -- & 3.15 \\
Latent (no comp.)        & 128 & 256 & 512 & 3.67 \\
Latent (comp. $r{=}8$)  & 16  & 32  & 64  & 0.46 \\
Latent (comp. $r{=}64$) & 2   & 4   & 8   & 0.057 \\
Latent (comp. $r{=}64$, no D3) & -- & 4 & 8 & 0.025 \\
\bottomrule
\end{tabular}
\vspace{1mm}

\footnotesize Channel counts per feature map; spatial resolutions:
$128{\times}128$ (D3), $64{\times}64$ (D4), and $32{\times}32$ (D5) for $1024{\times}1024$ input.
\vspace{-5mm}
\end{table}

\vspace{-2mm}
\section{Experiments}
\label{sec:experiments}
We experimentally evaluate the design choices introduced in Sec.~\ref{sec:method}, including early pre/post fusion, siamese processing, cross-attention mechanisms, and compression of pre-disaster latent representations. All configurations are trained and evaluated under a unified protocol, enabling controlled comparisons in terms of detection accuracy, robustness to misregistration, and system-level trade-offs.

\subsection{Dataset and Detection Setup}
\label{sec:dataset}
We conduct all experiments on the xBD dataset~\cite{gupta2019xbddatasetassessingbuilding}, introduced as part of the xView2 Challenge. The dataset provides pairs of coarsely registered pre- and post-event RGB satellite images for large-scale assessment of building damage following natural disasters.

Each sample consists of a pair of registered RGB images with a spatial resolution below  0.8\,m/px and a size of $1024 \times 1024$ pixels. Buildings are annotated with four discrete damage levels: \emph{no damage}, \emph{minor damage}, \emph{major damage}, and \emph{destroyed}. The dataset covers a wide range of disaster types, including earthquakes, tsunamis, volcanic eruptions, wildfires, wind-related events, and floods, resulting in significant variability in visual appearance and acquisition conditions.

Although xBD was originally designed for pixel-wise segmentation, we reformulate the task as object detection by converting building polygons provided in the GeoJSON annotations into axis-aligned bounding boxes. All evaluations are therefore performed in a detection setting.
 
Following the standard xBD split (excluding Tier~3 data), we use 4,665 image pairs partitioned into 60\% training, 20\% validation, and 20\% test sets. This fixed split is used consistently across all architectural variants to ensure fair comparison.

\subsection{Training and Evaluation Protocol}
\label{sec:protocol}
All experiments are conducted using a YOLOX-S backbone~\cite{ge2021yolox} unless explicitly stated otherwise. All architectural variants are trained independently under identical optimization settings to enable controlled comparisons.

\paragraph{Post-Disaster Shift Augmentation.}
To improve robustness to residual pre/post misregistration expected in onboard scenarios, we introduce a post-disaster shift augmentation (Aug) during training. A random spatial displacement is applied exclusively to the post-disaster image, with independent horizontal and vertical offsets uniformly sampled in $[0,150]$ pixels. Ground-truth annotations remain defined in the reference frame of the pre-disaster image. After shifting, only the spatial intersection between the pre- and post-disaster images is retained. Pixels outside this shared region and bounding boxes falling partially or fully outside it are discarded to preserve annotation consistency.

\paragraph{Evaluation and Test-Time Misregistration Protocol.}
Detection performance is evaluated using precision, recall, F1-score, and mAP@0.5. For robustness experiments, synthetic shifts are applied at test time to the post-disaster image only. Ground-truth annotations remain unchanged and are always defined in the reference frame of the pre-disaster image. To ensure fair comparison across shift magnitudes and directions, evaluation is performed on a fixed spatial support corresponding to the intersection induced by the maximum considered displacement. Pixels outside this shared region, as well as bounding boxes falling partially or entirely outside it, are removed consistently for all evaluated shifts. This guarantees that all robustness results are computed on an identical image subset and annotation set, independently of the applied displacement.

\subsection{Quantitative Results and Ablation Study} 
\label{sec:quantitative}

Tab.~\ref{tab:quantitative} reports quantitative results for the proposed architectural variants and reference baselines on the xBD dataset. We first compare detection-based approaches to segmentation-driven pipelines adapted for detection. The xView2 challenge winning solution, originally designed for semantic segmentation, exhibits substantially lower performance when adapted to object detection by extracting bounding boxes from segmentation masks. This performance gap should be interpreted with caution, as the original method was not optimized for instance-level detection, and the mask-to-bounding-box conversion can introduce additional localization errors, particularly for small or fragmented structures.

\begin{table}[t]
\centering
\caption{Quantitative comparison of architectural variants for on-board building damage detection on xBD dataset.}
\label{tab:quantitative}
\footnotesize
\setlength{\tabcolsep}{3pt}
\begin{tabular}{p{4.2cm}cccc}
\toprule
\textbf{Cfg.} & \textbf{P} & \textbf{R} & \textbf{F1} & \textbf{mAP@0.5} \\
\midrule
xView2 winner$^\dagger$ & 50.4 & 52.1 & 51.2 & 38.8 \\
\midrule
EF (6ch) & 58.3 & 57.0 & 57.4 & 57.4 \\
S & 61.1 & 57.8 & 59.1 & 58.2 \\
S + A & 62.4 & 54.0 & 57.5 & 57.1 \\
S + A$^{+D3}$ & 54.6 & 49.9 & 51.9 & 48.3 \\
\textbf{S + Aug} & \textbf{64.2} & \textbf{58.7} & \textbf{61.2} & \textbf{60.7} \\
S + A + Aug & 58.9 & 57.4 & 57.9 & 57.2 \\
\midrule
S + A + Aug + Comp(8) & 60.9 & 55.4 & 57.8 & 57.6 \\
S + A + Aug + Comp(64) & 59.8 & 55.6 & 57.4 & 57.0 \\
S + A + Aug + Comp(64)$-D3$ & 56.9 & 48.6 & 52.3 & 49.7 \\
\midrule
YOLOv12 S + A + Aug & 62.3 & 54.4 & 58.1 & 60.0 \\
\bottomrule
\end{tabular}
\vspace{1mm}

\begin{minipage}{\linewidth}
\footnotesize 
$^\dagger$~Seg → Det adaptation.
Acronyms:
\textbf{EF}: early fusion; 
\textbf{S}: siamese processing; 
\textbf{A}: cross-attention on D4--D5;
\textbf{A$^{+D3}$}: cross-attention extended to D3;
\textbf{Aug}: post-disaster shift augmentation during training;
\textbf{Comp($r$)}: channel-wise latent compression with ratio $r$; 
\textbf{$-D3$}: removal of the finest-resolution feature level from both branches.
\end{minipage}
\vspace{-5mm}
\end{table}

Using early fusion of pre- and post-disaster images as a baseline (EF), we observe that introducing siamese processing consistently improves performance, yielding a gain of approximately +1.5 points in F1-score and mAP. This confirms the benefit of explicitly separating pre- and post-disaster feature extraction prior to comparison, enabling more structured temporal reasoning.

Introducing cross-attention slightly decreases average detection performance in nominal conditions. In all reported results, attention is applied only at the coarser feature levels (D4--D5), as described in Sec.~\ref{sec:attention}. Extending attention to the finest-resolution feature level (D3) leads to a strong degradation (approximately -9\% mAP), supporting the design choice to exclude D3 from attention-based interactions.

From a training perspective, incorporating post-disaster shift augmentation (Aug) yields a significant improvement in average detection performance (+3\% mAP), indicating that exposure to synthetic misregistration enhances generalization even under nominal alignment. While architectures without cross-attention achieve the highest mean accuracy, attention-based variants become advantageous under large pre/post misregistrations, as analyzed in Sec.~\ref{sec:analysis}.

We further evaluate system-level constraints through compression of pre-disaster latent representations. Channel-wise compression with ratios up to $r=64$ induces only a marginal performance drop ($<1$\% mAP), while reducing the uplinked latent size from 3.7\,MB to 57\,kB (Tab.~\ref{tab:latent_compression}), corresponding to more than two orders of magnitude reduction. In contrast, removing the finest-resolution feature level (D3) results in a substantial accuracy degradation (approximately $-7$\% mAP), highlighting the importance of high-resolution features for precise building localization.

\begin{figure*}[ht]
\centering
\begin{subfigure}[t]{0.332\linewidth}
  \centering
  \includegraphics[width=\linewidth]{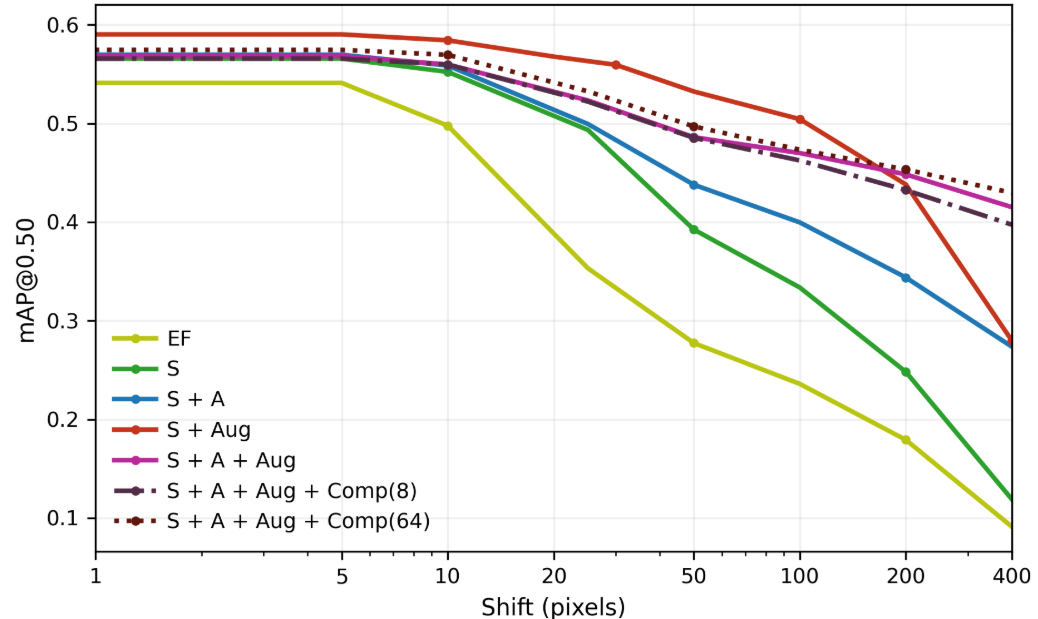}
  \caption{mAP@0.5 vs. post-image shift}
  \label{fig:shift_map}
\end{subfigure}\hfill
\begin{subfigure}[t]{0.3337\textwidth}
  \centering
  \includegraphics[width=\linewidth]{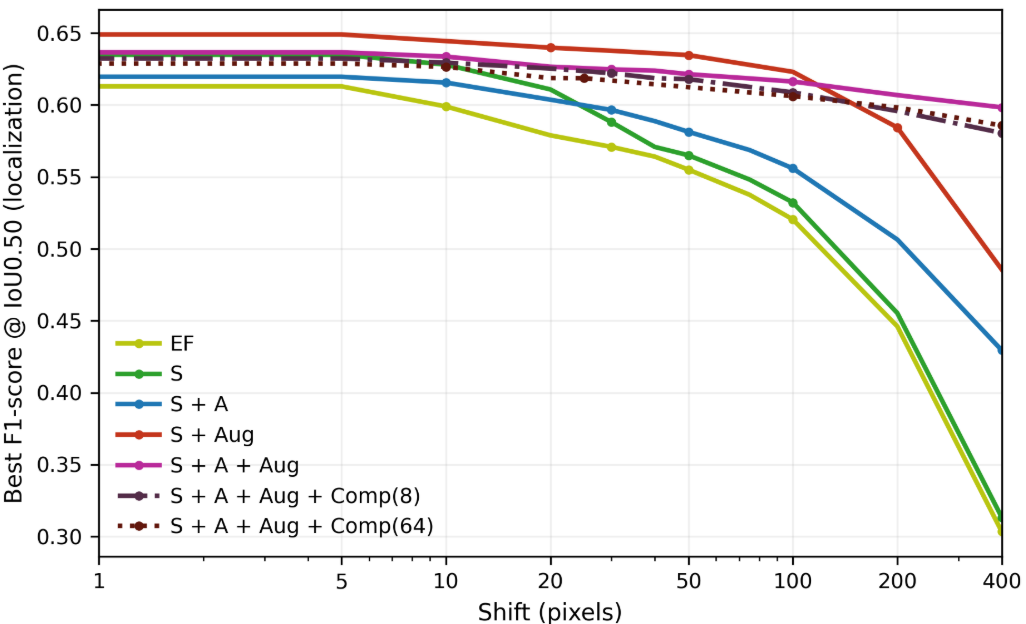}
  \caption{Localization F1-score vs. post-image shift}
  \label{fig:shift_loc}
\end{subfigure}\hfill
\begin{subfigure}[t]{0.334\textwidth}
  \centering
  \includegraphics[width=\linewidth]{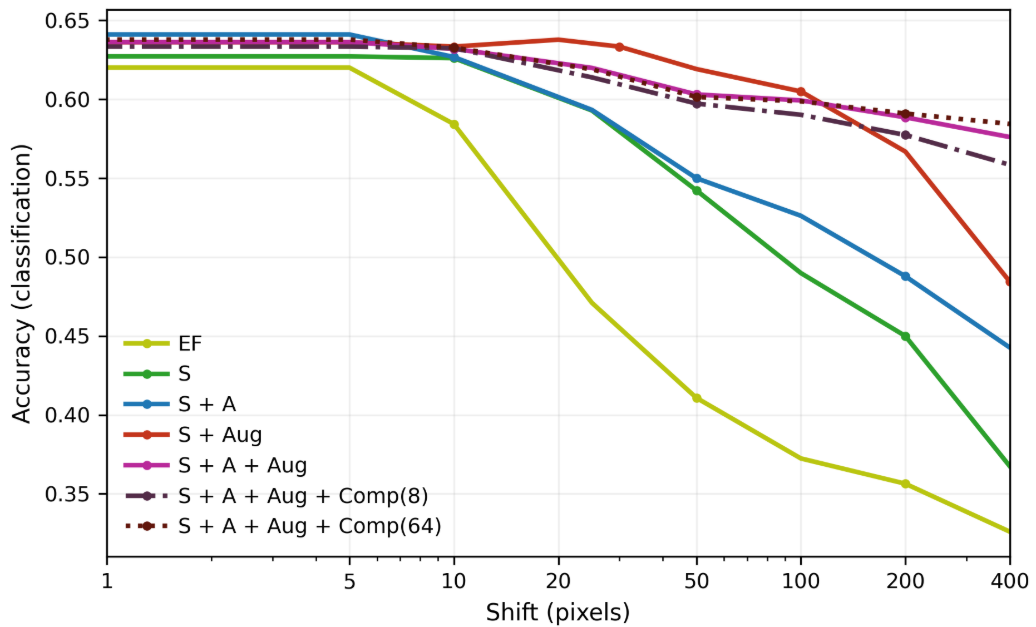}
  \caption{Classification accuracy vs. post-image shift}
  \label{fig:shift_cls}
\end{subfigure}
\caption{
Robustness to pre/post misregistration on xBD.
Performance as a function of the applied post-image shift magnitude.
Results are averaged over four shift directions. Legend:
\textbf{EF}: early fusion (6-channel input);
\textbf{S}: siamese processing;
\textbf{A}: cross-attention on D4--D5;
\textbf{Aug}: post-disaster shift augmentation during training;
\textbf{Comp($r$)}: channel-wise latent compression with ratio $r$.
}
\label{fig:robustness}
\vspace{-4mm}
\end{figure*}

Finally, we report results obtained with a more recent detection backbone (YOLOv12~\cite{tian2025yolov12attentioncentricrealtimeobject}) as a reference point. This experiment aims to position the proposed design choices relative to a newer generation of detectors that already incorporate attention mechanisms. While YOLOv12 yields a moderate performance improvement compared to the YOLOX-S baseline, the gain remains limited, and the relative trends observed across architectural variants remain consistent. This suggests that the conclusions drawn from our design space analysis are not tied to a specific detector implementation.

\subsection{On-board Robustness and Feasibility Analysis}
\label{sec:analysis}
We further analyze the proposed architectural variants along two complementary dimensions: robustness to pre/post image misregistration representative of on-board acquisition conditions, and system-level trade-offs related to communication, computation, and deployability on spaceborne platforms.

\subsubsection{Robustness to Pre/Post Misregistration}
\label{sec:misregistration}

In on-board EO scenarios, residual misregistration between pre- and post-disaster images is unavoidable due to acquisition geometry differences and limited co-registration capabilities. We evaluate robustness using the controlled test-time shift protocol described in Sec.~\ref{sec:protocol}, where synthetic displacements are applied to the post-disaster image while ground-truth annotations remain defined in the pre-disaster reference frame. Results are computed on the common spatial support defined by the maximum displacement.

Fig.~\ref{fig:shift_map} reports mAP@0.5 as a function of the misregistration magnitude. Architectures trained without Aug—whether early-fusion or siamese—exhibit a rapid performance drop for shifts on the order of a few tens of pixels, corresponding to typical building extents. In contrast, models trained with Aug show a markedly more gradual degradation. When the induced shift exceeds the network’s effective receptive field, performance eventually collapses, as reliable spatial correspondence between pre- and post-disaster features can no longer be established. Cross-attention further mitigates degradation under large misregistration when combined with Aug. Cross-attention alone (without Aug) provides limited robustness gains, indicating that the ability to retrieve spatially displaced features must be learned during training to be effective at inference. To better understand the origin of performance degradation, we separately analyze localization and damage classification. 

Localization performance (Fig.~\ref{fig:shift_loc}) is evaluated as building detection irrespective of damage level, using the maximum F1-score optimized over the object confidence threshold at IoU~$\geq$~0.5. A prediction is considered correct if it matches a ground-truth building with sufficient overlap, independently of its damage label. Localization degrades with increasing shift, but much less than damage classification and overall mAP. This relative stability stems from the fact that bounding boxes are predicted in the pre-disaster reference frame and mainly rely on features from that image. In particular, architectures combining Aug and cross-attention are almost insensitive to misregistration, suggesting that they have learned a robust localization behavior under spatial shifts.

Damage classification performance (Fig.~\ref{fig:shift_cls}) is evaluated only on correctly localized buildings and measured using standard classification accuracy over the four damage levels. In contrast to localization, damage classification is substantially more sensitive to misregistration and largely explains the observed mAP drop, since accurate class prediction requires consistent comparison between pre- and post-disaster features. Architectures combining Aug and cross-attention exhibit the slowest degradation in classification accuracy under large offsets, confirming their complementary roles. Importantly, channel-wise compression of pre-disaster latent representations does not modify the observed robustness trends for localization, classification, or overall mAP, indicating that aggressive data reduction remains compatible with misregistration tolerance.

\subsubsection{On-board Feasibility Analysis}
\label{sec:system}
Beyond detection accuracy and robustness, we assess the feasibility of deploying the proposed pipeline under realistic on-board constraints in terms of computation and communication. While the present study does not target a specific mission profile, we provide order-of-magnitude estimates to position the proposed approach with respect to existing spaceborne processing capabilities.

Previous works have demonstrated the deployment of YOLO-based object detection pipelines on embedded FPGA platforms equipped with dedicated AI accelerators, including recent Xilinx Versal devices~\cite{francesconievent,goudemant:hal-03881738}. Results reported on latest-generation embedded targets indicate processing throughputs on the order of $10^2$~MPixels/s for the neural inference stage alone, excluding image I/O and pre-processing. Assuming $1024\times1024$ image tiles at a ground sampling distance of approximately 0.8~m/px, this corresponds to processing roughly 100 tiles per second, i.e., about 70--80~km$^2$/s. Under these assumptions, a representative urban disaster area of 100~km$^2$ (roughly equivalent to the area of Paris proper) could be processed in approximately one to two seconds of pure inference time.

From a communication perspective, uplinking full-resolution pre-disaster images would require transmitting approximately 450~MB for a 100~km$^2$ area. In contrast, using the proposed compressed latent representation with a compression ratio of $r=64$ reduces the uplinked information to less than 10~MB for the same area, corresponding to more than two orders of magnitude reduction.

In addition, on-board processing enables drastic reduction of required downlink datarate by transmitting compact detection products (building bounding boxes and damage levels) instead of full-resolution imagery, with optional transmission of image patches limited to detected damaged regions. Overall, these estimates indicate that the proposed architecture offers a favorable trade-off between detection performance, robustness, and on-board resource usage.

\section{Discussion, Limitations, and Future Work}

Our experiments highlight several key advances enabling resource-efficient bi-temporal damage assessment. Siamese processing improves detection performance over early fusion while naturally supporting the separation between ground-encoded pre-disaster features and on-board post-disaster inference. Robustness-oriented shift augmentation—introduced to simulate realistic on-board misregistration—not only strengthens robustness to spatial offsets but also improves average detection accuracy under nominal conditions. Controlled shift experiments show that localization remains relatively stable under misalignment, whereas damage classification is more sensitive. Cross-attention helps maintain performance under large spatial shifts through non-local feature retrieval, while latent compression up to $r=64$ yields only marginal degradation and reduces uplink data volume by more than two orders of magnitude. Taken together, these findings represent a step toward the feasibility of bi-temporal object-level damage assessment under embedded inference and constrained ground-to-space bandwidth, using a ground/on-board latent encoding strategy, paving the way toward operational on-board deployment.

Several aspects must however be addressed to further increase system maturity. Current experiments rely on pre-processed imagery, whereas real on-board acquisitions involve raw data affected by sensor noise, radiometric variability, optical distortions, and residual geolocation errors \cite{Dorise}. Future work should evaluate robustness under realistic raw-data simulations and integrate lightweight preprocessing strategies. In addition, harmonization between pre- and post-event imagery remains challenging due to illumination and potential sensor differences; image uniformization strategies, combined with latent-space comparison, may help mitigate these discrepancies.

Beyond these maturity improvements, several exploratory directions can be considered. Leveraging multiple pre-disaster observations (reference time series) could produce more stable latent representations, improving localization robustness and reducing sensitivity to scene variability. Increasing the computational complexity of the ground segment—through larger encoders or additional contextual processing—while maintaining a lightweight on-board branch, offers a promising asymmetric architecture compatible with strict uplink budgets. Furthermore, although this study focuses on mono-modal optical imagery, the proposed ground/on-board architecture naturally extends to multi-sensor damage assessment. In operational contexts, heterogeneous observations (e.g., optical and Synthetic Aperture Radar) could enrich latent representations, although learning consistent cross-modal embeddings remains challenging and may require dedicated fusion strategies \cite{rs12020205}.

\section{Conclusion}
We introduced a unified ground/onboard architecture for object-level building damage assessment from bi-temporal satellite imagery, compatible with realistic satellite communication and compute constraints. The proposed approach works by first encoding, on ground, pre-disaster images into compact latent representations. These compressed representations are then used during post-disaster inference performed directly on board the satellite. By combining these two steps, this design enables efficient multi-temporal analysis close to the sensor while reducing the need to transmit large volumes of raw data.

Through a systematic design-space exploration, we showed that siamese processing and robustness-oriented data augmentation primarily improve average detection performance, while the combination of data augmentation and cross-attention mechanisms enhances robustness to severe pre/post misregistration. At the same time, aggressive latent-space compression achieves more than two orders of magnitude communication reduction with only marginal performance degradation, confirming the feasibility of embedded bi-temporal reasoning under stringent uplink constraints.

Beyond specific architectural choices, this work illustrates a shift in EO systems from ground-centric processing toward onboard applications that reduce communication load, improve responsiveness, and remain compatible with embedded hardware constraints.
{
    \small
    \bibliographystyle{ieeenat_fullname}
    \bibliography{main}
}


\end{document}